\def\year{2020}\relax
\documentclass[letterpaper,table]{article} 
\usepackage{aaai20}  
\usepackage{times}  
\usepackage{helvet} 
\usepackage{courier}  
\usepackage[hyphens]{url}  
\usepackage{graphicx} 
\usepackage{graphicx}  
\usepackage{pgfplots}
\pgfplotsset{compat=1.14}
\usepackage{tabu,multirow}
\usepackage{comment}
\usepackage[T1]{fontenc}
\usepackage[utf8]{inputenc}
\usepackage{dirtytalk}
\usepackage[T1]{fontenc}
\usepackage{color, colortbl}
\usepackage{subfig}
\usepackage{xcolor}
\usepackage{caption}
\usepackage{url}
\usepackage{csquotes}

\nocopyright
\title{\#MeTooMA: Multi-Aspect Annotations of Tweets Related to the MeToo Movement}
\author{Akash Gautam\thanks{equal contribution}$^{1}$, Puneet Mathur\footnotemark[1]$^{2}$, Rakesh Gosangi$^{3}$, Debanjan Mahata\thanks{Author participated in this research as an Adjunct Faculty at IIIT-Delhi.}$^{3}$,  \\ \bf \Large Ramit Sawhney$^{4}$, Rajiv Ratn Shah$^{1}$\\ 
$^{1}$ MIDAS, IIIT-Delhi \emph{\{akash15011, rajivratn\}@iiitd.ac.in}, \\
$^{2}$ University of Maryland, College Park \emph{puneetm@cs.umd.edu}, \\
$^{3}$ Bloomberg, New York, U.S.A. \emph{\{rgosangi, dmahata\}@bloomberg.net}, \\
$^{4}$ Netaji Subhas Institute of Technology \emph{ramits.co@nsit.net.in}\\
}

\begin{document}

\maketitle

\begin{abstract}
In this paper, we present a dataset containing 9,973 tweets related to the MeToo movement that were manually annotated for five different linguistic aspects: relevance, stance, hate speech, sarcasm, and dialogue acts. We present a detailed account of the data collection and annotation processes. The annotations have a very high inter-annotator agreement (0.79 to 0.93 k-alpha) due to the domain expertise of the annotators and clear annotation instructions. We analyze the data in terms of geographical distribution, label correlations, and keywords. Lastly, we present some potential use cases of this dataset. We expect this dataset would be of great interest to psycholinguists, socio-linguists, and computational linguists to study the discursive space of digitally mobilized social movements on sensitive issues like sexual harassment. 

\end{abstract}

\section*{Introduction}
Over the last couple of years, the MeToo movement has facilitated several discussions about sexual abuse. Social media, especially Twitter, was one of the leading platforms where people shared their experiences of sexual harassment, expressed their opinions, and also offered support to victims. A large portion of these tweets was tagged with a dedicated hashtag \#MeToo, and it was one of the main trending topics in many countries. The movement was viral on social media and the hashtag used over 19 million times\footnote{https://www.usatoday.com/story/news/2018/10/13/metoo-impact-hashtag-made-online/1633570002/} in a year. 

The MeToo movement has been described as an essential development against the culture of sexual misconduct by many feminists, activists, and politicians. It is one of the primary examples of successful digital activism facilitated by social media platforms. The movement generated many conversations on stigmatized issues like sexual abuse and violence, which were not often discussed before because of the associated fear of shame or retaliation. This creates an opportunity for researchers to study how people express their opinion on a sensitive topic in an informal setting like social media. However, this is only possible if there are annotated datasets that explore different linguistic facets of such social media narratives. 

Twitter served as a platform for many different types of narratives during the MeToo movement \cite{hosterman2018twitter}. It was used for sharing personal stories of abuse, offering support and resources to victims, and expressing support or opposition towards the movement \cite{lopez2019one}. It was also used to allege individuals of sexual misconduct, refute such claims, and sometimes voice hateful or sarcastic comments about the campaign or individuals. In some cases, people also misused hashtag to share irrelevant or uninformative content. To capture all these complex narratives, we decided to curate a dataset of tweets related to the MeToo movement that is annotated for various linguistic aspects.

In this paper, we present a new dataset (MeTooMA\footnote{The dataset can be found at \url{ https://doi.org/10.7910/DVN/JN4EYU}.}) that contains 9,973 tweets associated with the MeToo movement annotated for relevance, stance, hate speech, sarcasm, and dialogue acts. We introduce and annotate three new dialogue acts that are specific to the movement: Allegation, Refutation, and Justification. The dataset also contains geographical information about the tweets: from which country it was posted. 

We expect this dataset would be of great interest and use to both computational and socio-linguists. For computational linguists, it provides an opportunity to model three new complex dialogue acts (allegation, refutation, and justification) and also to study how these acts interact with some of the other linguistic components like stance, hate, and sarcasm. For socio-linguists, it provides an opportunity to explore how a movement manifests in social media across multiple countries.  

\definecolor{Gray}{gray}{0.9}

\begin{table*}
\centering
\begin{tabular}{|l|l|l|}
\hline
\textbf{Dataset} & \textbf{\#Annotated Posts} &\parbox[s]{3.5cm}{}{\textbf{Labels}} \\
\rowcolor{Gray}
\cite{pandey2018distributional} &2500 & \textit{accusational, validation, sensational}  \\
\cite{khatua2018sounds} &1024 & \parbox{6.5cm}{\textit{assault at: workplace, educational institute, public place, home}}     \\
\rowcolor{Gray}
\cite{schrading2015analysis} &18,336 & {\textit{abuse, non-abuse}} \\

\cite{chowdhury2019speak} &5119 &\textit{recollection, non-recollection} \\
\rowcolor{Gray}
\cite{sharifirad2019learning} &3240 & {\textit{indirect, sexism, casual sexism, physical sexism}} \\
MeTooMA & 9,937 & \parbox{6.5cm}{\textit{relevance, stance, hate speech, sarcasm, dialogue acts (allegation, justification, refutation)}} \\

\hline
\end{tabular}
\caption{\textbf{Summary of related datasets.}}
\label{tab: summary}
\end{table*}

\section*{Related Datasets}
\label{sec:rel_work}
Table \ref{tab: summary} presents a summary of datasets that contain social media posts about sexual abuse and annotated for various labels. 
\begin{itemize}
     \item \cite{pandey2018distributional} created a dataset of 2,500 tweets for identification of malicious intent surrounding the cases of sexual assault. The tweets were annotated for labels like \textit{accusational, validation, sensational}.
     
     \item Khatua \emph{et al} \cite{khatua2018sounds} collected  0.7 million tweets containing hashtags such as \textit{\#MeToo}, \textit{\#AlyssaMilano}, \textit{\#harassed}. The annotated a subset of 1024 tweets for the following assault-related labels: assault at the workplace by colleagues, assault at the educational institute by teachers or classmates, assault at public places by strangers, assault at home by a family member, multiple instances of assaults, or a generic tweet about sexual violence. 
     
     \item \cite{schrading2015analysis} created the Reddit Domestic Abuse Dataset, which contained 18,336 posts annotated for 2 classes, \textit{abuse} and \textit{non-abuse}.
     
     \item \cite{chowdhury2019speak} presented a dataset consisting of 5119 tweets distributed into \textit{recollection} and \textit{non-recollection} classes. The tweet was annotated as \textit{recollection} if it explicitly mentioned a personal instance of sexual harassment. 
     
     \item Sharifirad \emph{et al} \cite{sharifirad2019learning} created a dataset with 3240 tweets labeled into three categories of sexism: \textit{Indirect sexism, casual sexism, physical sexism}. 
      
\end{itemize}

\noindent SVAC (Sexual Violence in Armed Conflict) is another related dataset which contains reports annotated for six different aspects of sexual violence: \textit{prevalence}, \textit{perpetrators}, \textit{victims}, \textit{forms}, \textit{location}, and \textit{timing}.

Unlike all the datasets described above, which are annotated for a single group of labels, our dataset is annotated for \textbf{five different linguistic aspects}. It also has \textbf{more annotated samples} than most of its contemporaries.

\section*{Dataset}
\label{sec:dataset}

\subsection{Data Collection}
We focused our data collection over the period of October to December 2018 because October marked the one year anniversary of the MeToo movement. Our first step was to identify a list of countries where the movement was trending during the data collection period. To this end, we used Google's interactive tool named MeTooRisingWithGoogle\footnote{https://metoorising.withgoogle.com/}, which visualizes search trends of the term "MeToo" across the globe. This helped us narrow down our query space to 16 countries. 

We then scraped 500 random posts from online sexual harassment support forums to help identify keywords or phrases related to the movement \footnote{We scraped data from the discussion forums on the websites of two non-profit organizations (pandys and isurvive), which provide support and resources to survivors of abuse.}. The posts were first manually inspected by the annotators to determine if they were related to the MeToo movement. Namely, if they contained self-disclosures of sexual violence, relevant information about the events associated with the movement, references to news articles or advertisements calling for support for the movement. We then processed the relevant posts to extract a set of uni-grams and bi-grams with high tf-idf scores. The annotators further pruned this set by removing irrelevant terms resulting in a lexicon of 75 keywords. Some examples include: \#Sexual Harassment, \#TimesUp, \#EveryDaySexism, assaulted, \#WhenIwas, inappropriate, workplace harassment, groped, \#NotOkay, believe survivors, \#WhyIDidntReport.

We then used Twitter's public streaming API\footnote{https://www.tweepy.org/} to query for tweets from the selected countries, over the chosen three-month time frame, containing any of the keywords. This resulted in a preliminary corpus of 39,406 tweets. We further filtered this data down to include only English tweets based on tweet's \textit{language} metadata field and also excluded short tweets (less than two tokens). Lastly, we de-duplicated the dataset based on the textual content. Namely, we removed all tweets that had more than 0.8 cosine similarity score on the unaltered text in tf-idf space with any another tweet. We employed this de-duplication to promote more lexical diversity in the dataset. After this filtering, we ended up with a corpus of 9,973 tweets. 

\begin{figure}[t]
\centering
\includegraphics[scale=0.363, height=3.6cm]{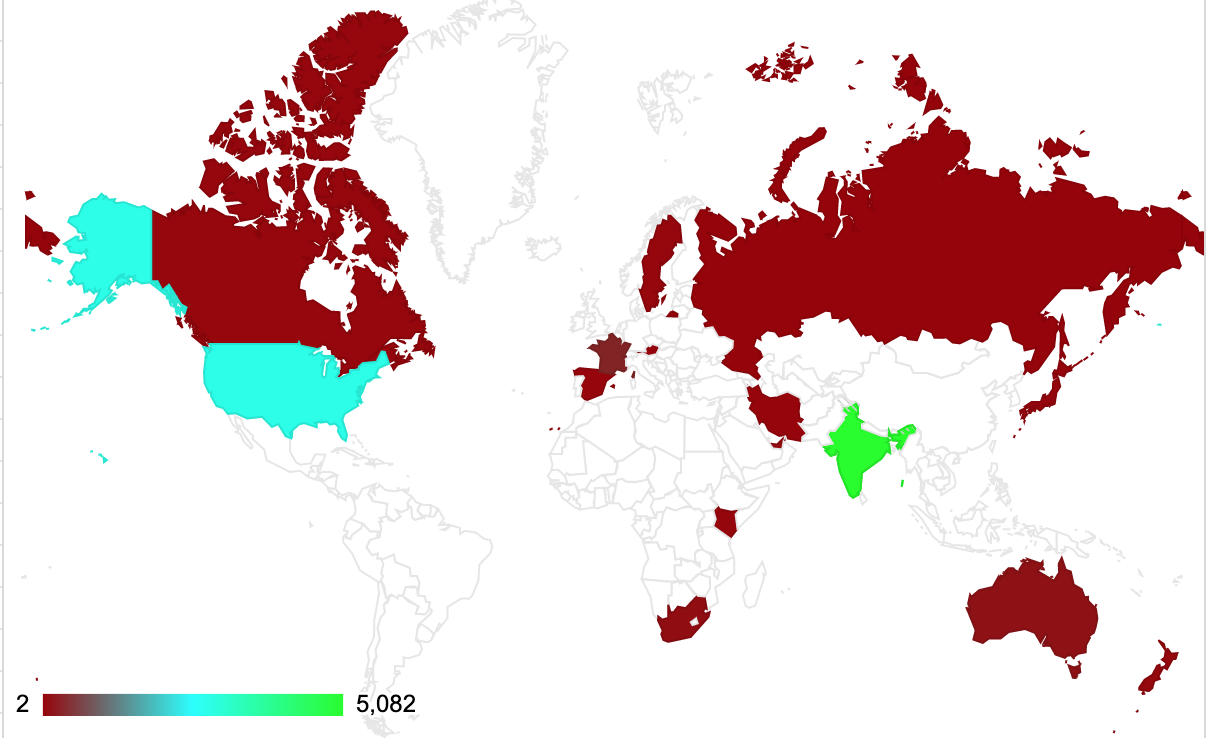}
\caption{\textbf{Choropleth world map recording tweet frequency.}}
\label{fig:heatmap}
\end{figure}

Table \ref{tab: count_tweet_real} presents the distribution of the tweets by country before and after the filtering process. A large portion of the samples is from India because the MeToo movement has peaked towards the end of 2018 in India. There are very few samples from Russia likely because of content moderation and regulations on social media usage in the country\footnote{https://time.com/5636107/metoo-russia-womens-rights/}. Figure \ref{fig:heatmap} gives a geographical distribution of the curated dataset.

\textbf{\textit{Due to the sensitive nature of this data, we have decided to remove any personal identifiers (such as names, locations, and hyperlinks) from the examples presented in this paper. We also want to caution the readers that some of the examples in the rest of the paper, though censored for profanity, contain offensive language and express a harsh sentiment.}}

\begin{table}[]
\centering
\begin{tabular}{|lll|}
\hline
\textbf{Country} & \textbf{\#Tweets} &\parbox{2.5cm}{\textbf{\#Filtered Tweets}} \\ 
\hline
India   &20,112   &5,082     \\ 
\rowcolor{Gray}
USA        &8,943  &2,773      \\ 
United Kingdom  &4,350 &1,334 \\
\rowcolor{Gray}
France &1,120 &347        \\ 
Australia &542 &153     \\
\rowcolor{Gray}
South Africa &1,085 &103  \\ 
Japan &830 &13   \\ 
\rowcolor{Gray}
Kenya &696 &15  \\ 
UAE &540 &51        \\
\rowcolor{Gray}
New Zealand &248 &38    \\
Iran &325 &7 \\ 
\rowcolor{Gray}
Canada &324 &24 \\ 
Sweden &139 &20 \\ 
\rowcolor{Gray}
Spain &62 &9 \\
Austria &88 &2    \\ 
\rowcolor{Gray}
Russia &42 &2  \\ \hline
\textbf{Total} & \textbf{39,406}  &\textbf{9,973}  \\ \hline    
\end{tabular}
\caption{\textbf{Distribution of tweets by the country.}}
\label{tab: count_tweet_real}
\end{table}

\subsection{Annotation Task}
We chose against crowd-sourcing the annotation process because of the sensitive nature of the data and also to ensure a high quality of annotations. We employed three domain experts who had advanced degrees in clinical psychology and gender studies. The annotators were first provided with the guidelines\footnote{The annotation guidelines will be released as supplementary material to this publication.} document, which included instructions about each task, definitions of class labels, and examples. They studied this document and worked on a few examples to familiarize themselves with the annotation task. They also provided feedback on the document, which helped us refine the instructions and class definitions. The annotation process was broken down into five sub-tasks: for a given tweet, the annotators were instructed to identify relevance, stance, hate speech, sarcasm, and dialogue act. An important consideration was that the sub-tasks were not mutually exclusive, implying that the presence of one label did not consequently mean an absence of any.

\subsubsection{Task 1: Relevance} 
Here the annotators had to determine if the given tweet was relevant to the MeToo movement. Relevant tweets typically include personal opinions (either positive or negative), experiences of abuse, support for victims, or links to MeToo related news articles. 
Following are examples of a \textit{relevant} tweet:
\begin{quote}
\small
    \textit{Officer [name] could be kicked out of the force after admitting he groped a woman at [place] festival last year. His lawyer argued saying the constable shouldn't be punished because of the \#MeToo movement. \#notokay \#sexualabuse.}
\end{quote}
and an \textit{irrelevant} tweet:
\begin{quote}
\small
    \textit{Had a bit of break. Went to the beautiful Port [place] and nearby areas. Absolutely stunning as usual. \#beautiful \#MeToo \#Australia \#auspol [URL].}
\end{quote}
We expect this relevance annotation could serve as a useful filter for downstream modeling. 

\subsubsection{Task 2: Stance}
Stance detection is the task of determining if the author of a text is in favour or opposition of a particular target of interest \cite{augenstein2016stance,mohammad2016semeval}. Stance helps understand public opinion about a topic and also has downstream applications in information extraction, text summarization, and textual entailment \cite{sobhani2017stance}. We categorized stance into three classes: Support, Opposition, Neither. Support typically included tweets that expressed appreciation of the MeToo movement, shared resources for victims of sexual abuse, or offered empathy towards victims. Following is an example of a tweet with a \textit{Support} stance:

\begin{quote}
    \small
    \textit{Opinion: \#MeToo gives a voice to victims while bringing attention to a nationwide stigma surrounding sexual misconduct at a local level.[URL]. This should go on.}
\end{quote}

\noindent On the other hand, Opposition included tweets expressing dissent over the movement or demonstrating indifference towards the victims of sexual abuse or sexual violence. An example of an \textit{Opposition} tweet is shown below:
\begin{quote}
    \small
    \textit{The double standards and selective outrage make it clear that feminist concerns about power imbalances in the workplace aren't principles but are tools to use against powerful men they hate and wish to destroy. \#fakefeminism. \#men.}
\end{quote}

\subsubsection{Task 3: Hate Speech}
Detection of hate speech in social media has been gaining interest from NLP researchers lately \cite{waseem2016hateful,badjatiya2017deep}. Our annotation scheme for hate speech is based on the work of \cite{basile-etal-2019-semeval}. For a given tweet, the annotators first had to determine if it contained any hate speech. If the tweet was hateful, they had to identify if the hate was \textit{Directed} or \textit{Generalized}. Directed hate is targeted at a particular individual or entity, whereas Generalized hate is targeted at larger groups that belonged to a particular ethnicity, gender, or sexual orientation. Following are examples of tweets with \textit{Directed} hate:
\begin{quote}
\small
\textit{[username] were lit minus getting f*c*i*g mouthraped by some drunk chick \#MeToo (no body cares because I'm a male) [URL]}
\end{quote}
and \textit{Generalized} hate:
\begin{quote}
\small
\textit{For the men who r asking "y not then, y now?", u guys will still doubt her \& harrass her even more for y she shared her story immediately no matter what! When your sister will tell her childhood story to u one day, i challenge u guys to ask "y not then, y now?" \#Metoo [username] [URL] \#a**holes.}
\end{quote}

\subsubsection{Task 4: Sarcasm}
Sarcasm detection has also become a topic of interest for computational linguistics over the last few years \cite{bamman2015contextualized,rajadesingan2015sarcasm} with applications in areas like sentiment analysis and affective computing. Sarcasm was an integral part of the MeToo movement. For example, many women used the hashtag \#NoWomanEver to sarcastically describe some of their experiences with harassment\footnote{https://www.good.is/articles/maura-quint-twitter-sexual-assault}. We instructed the annotators to identify the presence of any sarcasm in a tweet either about the movement or about an individual or entity. Following is an example of a sarcastic tweet:
\begin{quote}
\small
\textit{\# was pound before it was a hashtag. If you replace hashtag with the pound in the \#metoo, you get pound me too. Does that apply to [name].}
\end{quote}

\subsubsection{Task 5: Dialogue Acts}
A dialogue act is defined as the function of a speaker's utterance during a conversation \cite{mctear2016conversational}, for example, question, answer, request, suggestion, etc. Dialogue Acts have been extensive studied in spoken \cite{ang2005automatic} and written \cite{kim2010classifying} conversations and have lately been gaining interest in social media \cite{zarisheva2015dialog}. In this task, we introduced three new dialogue acts that are specific to the MeToo movement: Allegation, Refutation, and Justification. 

\textbf{Allegation}: This category includes tweets that allege an individual or a group of sexual misconduct. The tweet could either be personal opinion or text summarizing allegations made against someone \cite{hutchings2012commercial}. The annotators were instructed to identify if the tweet includes the hypothesis of allegation based on first-hand account or a verifiable source confirming the allegation. Following is an example of a tweet that qualifies as an Allegation: 

\begin{quote}
\small
\textit{More women accuse [name] of grave sexual misconduct...twitter seethes with anger. \#MeToo \#pervert.}
\end{quote}

\textbf{Refutation}: This category contains tweets where an individual or an organization is denying allegations with or without evidence. Following is an example of a Refutation tweet:

\begin{quote}
\small
\textit{She is trying to use the \#MeToo movement to settle old scores, says [name1] after [name2] levels sexual assault allegations against him.}
\end{quote}

\textbf{Justification}: The class includes tweets where the author is justifying their actions. These could be alleged actions in the real world (e.g. allegation of sexual misconduct) or some action performed on twitter (e.g. supporting someone who was alleged of misconduct). Following is an example of a tweet that would be tagged as Justification: 
\begin{quote}
\small
\textit{I actually did try to report it, but he and of his friends got together and lied to the police about it. \#WhyIDidNotReport.}
\end{quote}

\begin{figure}
\centering
\includegraphics[scale=0.415]{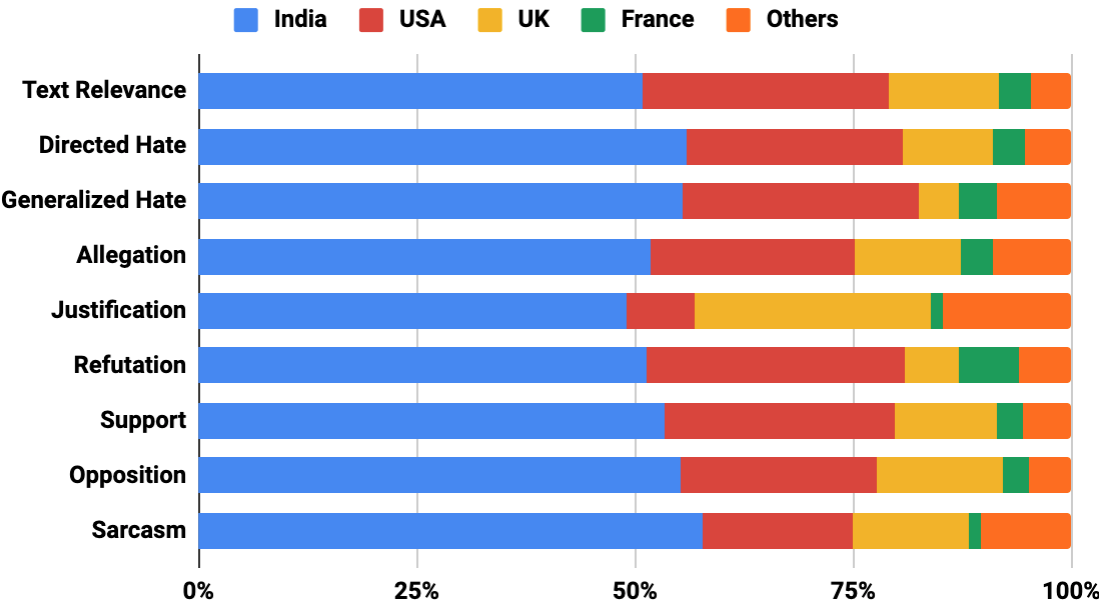}
\caption{\textbf{Geographical distribution of various class labels.}}
\label{fig:mannual_annotation}
\end{figure}

\newcommand\crule[3][black]{\textcolor{#1}{\rule{#2}{#3}}}
\begin{table}[ht]
\centering
\begin{tabular}{|llll|}
\hline
\rowcolor{Gray}
    \parbox{1.7cm}{\textbf{Directed Hate}} &\textbf{SAGE} &\parbox{1.7cm}{\textbf{Generalized Hate}} &\textbf{SAGE} \\
    
    f*ck&\crule[red!50!white!100]{6.36mm}{0.25cm}3.36  &hate &\crule[red!50!white!100]{6.21mm}{0.25cm}3.21\\
     \rowcolor{Gray}
     f*cking &\crule[red!50!white!100]{6.04mm}{0.25cm}3.04 &lie &\crule[red!50!white!100]{5.95mm}{0.25cm}2.95   \\
     hijab &\crule[red!50!white!100]{5.84mm}{0.25cm}2.84 &predators &\crule[red!50!white!100]{5.92mm}{0.25cm}2.92 \\
     \rowcolor{Gray}
     bullshit &\crule[red!50!white!100]{5.77mm}{0.25cm}2.77 &nuns &\crule[red!50!white!100]{5.91mm}{0.25cm}2.91 \\
     blog &\crule[red!50!white!100]{5.70mm}{0.25cm}2.70 &grop &\crule[red!50!white!100]{5.91mm}{0.25cm}2.91 \\
     \rowcolor{Gray}
     \parbox{1.2cm}{\textbf{Allegation}} &\textbf{SAGE} &\parbox{1.5cm}{\textbf{Justification}} &\textbf{SAGE} \\
     accuse &\crule[red!50!white!100]{4.45mm}{0.25cm}1.45 & organisation &\crule[red!50!white!100]{3.57mm}{0.25cm}0.57      \\
     \rowcolor{Gray}
     bob &\crule[red!50!white!100]{4.91mm}{0.25cm}1.45 & told &\crule[red!50!white!100]{3.91mm}{0.25cm}0.56                 \\
     flopping &\crule[red!50!white!100]{4.40mm}{0.25cm}1.40 &discuss &\crule[red!50!white!100]{3.91mm}{0.25cm}0.56          \\
     \rowcolor{Gray}
     aces &\crule[red!50!white!100]{4.40mm}{0.25cm}1.40 &violent &\crule[red!50!white!100]{3.55mm}{0.25cm}0.55              \\
     corrupt &\crule[red!50!white!100]{4.35mm}{0.25cm}1.35 &shocked &\crule[red!50!white!100]{3.51mm}{0.25cm}0.51           \\
     \rowcolor{Gray}
     
     \parbox{1.2cm}{\textbf{Support}} &\textbf{SAGE} &\parbox{1.5cm}{\textbf{Opposition}} &\textbf{SAGE} \\
         
    fund & \crule[red!50!white!100]{3.80mm}{0.25cm}0.80 &mocks &\crule[red!50!white!100]{5.47mm}{0.25cm}2.47  \\
    \rowcolor{Gray}
    reconciliation &\crule[red!50!white!100]{3.66mm}{0.25cm}0.66 &tweet &\crule[red!50!white!100]{5.19mm}{0.25cm}2.19   \\
    diversity &\crule[red!50!white!100]{3.62mm}{0.25cm}0.62 &practice &\crule[red!50!white!100]{5.19mm}{0.25cm}2.19     \\
    \rowcolor{Gray}
    protect &\crule[red!50!white!100]{3.62mm}{0.25cm}0.62 &feminism &\crule[red!50!white!100]{5.11mm}{0.25cm}2.11       \\
    welcome &\crule[red!50!white!100]{3.59mm}{0.25cm}0.59 &minister &\crule[red!50!white!100]{5.11mm}{0.25cm}2.11       \\
    \rowcolor{Gray}
    
     \parbox{1.2cm}{\textbf{Refutation}} &\textbf{SAGE} &\parbox{1.5cm}{\textbf{Sarcasm}} &\textbf{SAGE} \\
     
     baseless &\crule[red!50!white!100]{6.63mm}{0.25cm}3.63 & lol &\crule[red!50!white!100]{5.74mm}{0.25cm}2.74 \\
     \rowcolor{Gray}
     wild &\crule[red!50!white!100]{6.59mm}{0.25cm}3.59 &gonna &\crule[red!50!white!100]{5.71mm}{0.25cm}2.71    \\
     center &\crule[red!50!white!100]{6.46mm}{0.25cm}3.46 &trouble &\crule[red!50!white!100]{5.71mm}{0.25cm}2.71 \\
     \rowcolor{Gray}
     denies &\crule[red!50!white!100]{6.17mm}{0.25cm}3.17 &ooh &\crule[red!50!white!100]{5.41mm}{0.25cm}2.41    \\
     threatens &\crule[red!50!white!100]{6.07mm}{0.25cm}3.07 &xoxo &\crule[red!50!white!100]{5.20mm}{0.25cm}2.20 \\
     
\hline     
\end{tabular}
\caption{\textbf{Top five phrases learned by SAGE Topic model for the all the labels}}
\label{tab:sage_res}
\end{table}

\section*{Dataset Analysis}
\label{sec:description}
This section includes descriptive and quantitative analysis performed on the dataset. 

\subsection{Inter-annotator agreement}
We evaluated inter-annotator agreements using Krippendorff's alpha (K-alpha) \cite{krippendorff2011computing}. K-alpha, unlike simple agreement measures, accounts for chance correction and class distributions and can be generalized to multiple annotators. Table \ref{tab:aggreeemnts} summarizes the K-alpha measures for all the annotation tasks. We observe very strong agreements for most of the tasks with a maximum of 0.92 for the relevance task. The least agreement observed was for the hate speech task at 0.78. Per recommendations in \cite{artstein2008inter}, we conclude that these annotations are of good quality. We chose a straightforward approach of majority decision for label adjudication: if two or more annotators agreed on assigning a particular class label. In cases of discrepancy, the labels were adjudicated manually by the authors. Table \ref{tab:class_dist} shows a distribution of class labels after adjudication.

\begin{table}[]
\centering
\begin{tabular}{|ll|}
\hline
\textbf{Task} &\textbf{Krippendorff's $\alpha$}\\
\hline
\rowcolor{Gray}
Relevance & 0.92 \\
Stance & 0.90 \\
\rowcolor{Gray}
Hate speech & 0.78 \\
Sarcasm & 0.80 \\
\rowcolor{Gray}
Allegation & 0.86 \\
Refutation & 0.83 \\
\rowcolor{Gray}
Justification & 0.79 \\
\hline
\end{tabular}
\caption{\textbf{Inter-annotator agreements for all the annotation tasks.}}
\label{tab:aggreeemnts}
\end{table}

\begin{table}[]
\centering
\begin{tabular}{|llll|}
\hline
\textbf{Task} & \textbf{Label} & \textbf{\#Samples} & \textbf{\%}\\
\hline
\rowcolor{Gray}
Relevance & Relevant & 7,249 & 72.8\% \\
Stance & Support & 3,074 & 30.9\% \\
    & Opposition & 743 & 7.4\% \\
\rowcolor{Gray}    
Hate Speech & Directed & 419 & 4.21\% \\
\rowcolor{Gray}
            & Generalized & 281 & 2.8\% \\
Sarcasm & Sarcastic & 220 & 2.2\% \\
\rowcolor{Gray}
Dialogue Acts & Allegation & 578 & 5.78\% \\
\rowcolor{Gray}
& Justification & 292 & 2.9\% \\
\rowcolor{Gray}
& Refutation & 216 & 2.1\% \\
\hline
\end{tabular}
\caption{\textbf{Distribution of class labels for all tasks.}}
\label{tab:class_dist}
\end{table}

\begin{figure}
\centering
\includegraphics[scale=0.413]{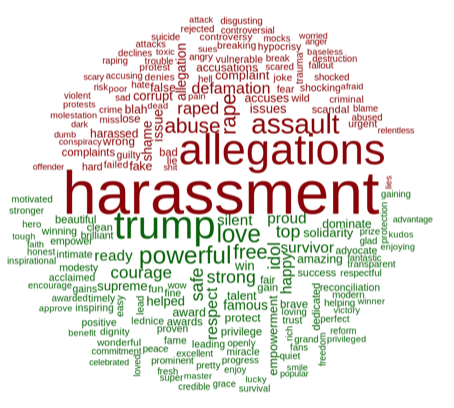}
\caption{\textbf{Word cloud representation of the dataset: font size is proportional to the frequency of a term. The words are organized and color-coded based on the NRC sentiment lexicon: positive sentiment (green + bottom half), negative sentiment (red + top half).}}
\label{fig:nrc_sentiment}
\end{figure}

\subsection{Geographical Distribution}
Figure \ref{fig:mannual_annotation} presents a distribution of all the tweets by their country of origin. As expected, a large portion of the tweets across all classes are from India, which is consistent with Table \ref{tab: count_tweet_real}. Interestingly, the US contributes comparatively a smaller proportion of tweets to Justification category, and likewise, UK contributes a lower portion of tweets to the Generalized Hate category. Further analysis is necessary to establish if these observations are statistically significant. 

\subsection{Label Correlations}
We conducted a simple experiment to understand the linguistic similarities (or lack thereof) for different pairs of class labels both within and across tasks. To this end, for each pair of labels, we converted the data into its tf-idf representation and then estimated Pearson, Spearman, and Kendall Tau correlation coefficients and also the corresponding $p$ values. The results are summarized in Table \ref{tab:corr_coeff}. Overall, the correlation values seem to be on a lower end with maximum Pearson's correlation value obtained for the label pair \textit{Justification - Support}, maximum Kendall Tau's correlation for \textit{Allegation - Support}, and maximum Spearman's correlation for \textit{Directed Hate - Generalized Hate}. The correlations are statistically significant ($p$ $<$ 0.05) for three pairs of class labels: \textit{Directed Hate - Generalized Hate, Directed Hate - Opposition, Sarcasm - Opposition}. Sarcasm and Allegation also have statistically significant $p$ values for Pearson and Spearman correlations.

\begin{table}[!ht]
\centering
\resizebox{\columnwidth}{!}{\begin{tabular}{|lllllll|}
\hline
\textbf{Label pair}                       & \textbf{PCC}   & \textbf{p-PCC} & \textbf{KCC}   & \textbf{p-KCC} & \textbf{SCC}    & \textbf{p-SCC}   \\
\hline
\textbf{Directed Hate - Generalized Hate} & 0.049 &   \textbf{0.0432}    & 0.268 & \textbf{0.0021}      & \textbf{0.477}  &  \textbf{0.0344}\\
\rowcolor{Gray}
Directed Hate - Sarcasm          & 0.052 &   0.0731    & 0.252 &   0.0521    & 0.258  & 0.0623        \\
Directed Hate - Allegation        & 0.045  &  0.0832     & 0.244 &  0.0712     & 0.252  &   0.0523      \\
\rowcolor{Gray}
Directed Hate - Justification      & 0.049  & 0.0661      & 0.413  &  0.0053     & 0.381   & 0.0503        \\
Directed Hate - Refutation         & 0.054  & 0.5391    & 0.314 &   0.0044    & 0.322  & 0.0712     \\
\rowcolor{Gray}
Directed Hate - Support            & 0.073   & 0.0882      & 0.042  &  0.0621     & 0.303   & 0.0032        \\
\textbf{Directed Hate - Opposition}             & 0.061  & \textbf{0.0022}      & 0.314 &   \textbf{0.0450}    & 0.322  &   \textbf{0.0433}      \\
\rowcolor{Gray}
Generalized Hate - Sarcasm        & 0.062  & 0.0233   & 0.260 &  0.0051     & 0.265  &   0.0421      \\
Generalized Hate - Allegation           & 0.059  & 0.0644      & 0.266 &  0.0260     & 0.271   &   0.0345      \\
\rowcolor{Gray}
Generalized Hate - Justification        & 0.034  & 0.0633      & 0.271  &  0.0532     & 0.281  &    0.0611     \\
Generalized Hate -Refutation            & 0.051  & 0.0821      & 0.223  &  0.0558     & 0.230     & 0.0031        \\
\rowcolor{Gray}
Generalized Hate - Support              & 0.028  &  0.6820    & 0.325  &    0.0621   & 0.355   & 0.0652       \\
Generalized Hate - Opposition               & 0.068  & 0.0239      & 0.320  &  0.0030     & 0.341   &  0.0532       \\
\rowcolor{Gray}
Sarcasm - Allegation               & 0.045  & 0.0471     & 0.244 & 0.0613 & 0.202 & 0.0072 \\
Sarcasm - Justification          & 0.061  & 0.0891     & 0.281  & 0.0401  & 0.013   & 0.0014    \\
\rowcolor{Gray}
Sarcasm - Refutation                & 0.035  & 0.0772     & 0.243  & 0.0023 & 0.221   & 0.0833        \\
Sarcasm - Support                  & 0.064  & 0.0514     & 0.233 & 0.0080 & 0.259  & 0.0041    \\
\rowcolor{Gray}
\textbf{Sarcasm - Opposition}
& 0.062  & \textbf{0.0034}     & 0.271  & \textbf{0.0430} & 0.362   & \textbf{0.0332}    \\
Allegation - Justification         & 0.053  & 0.0499     & 0.251  & 0.0031  & 0.262   & 0.0023    \\
\rowcolor{Gray}
Allegation - Refutation             & 0.062  & 0.0344     & 0.280  & 0.0421  & 0.281   & 0.0014    \\
Allegation - Support               & 0.027  & 0.6711      & \textbf{0.467}  &  0.0631     & 0.003   &  0.0779       \\
\rowcolor{Gray}
Allegation - Opposition                & 0.574  & 0.6533      & 0.359  & 0.0231 & 0.205   & 0.0702   \\
Justification - Refutation          & 0.443  & 0.6688      & 0.226  & 0.0711  & 0.226  & 0.0244    \\
\rowcolor{Gray}
Justification - Support              & \textbf{0.742}  & 0.7121      & 0.311  &  0.0093     & 0.311    & 0.0261        \\
Justification - Opposition             & 0.734   &   0.0429    & 0.326  & 0.0201     & 0.385   &  0.0342       \\
\rowcolor{Gray}
Refutation - Support               & 0.562  &   0.0822    & 0.237 & 0.0718  & 0.252   & 0.0522    \\
Refutation - Opposition                 & 0.651  &  0.0633     & 0.433   & 0.0433      & 0.043   & 0.0521        \\
\rowcolor{Gray}
Support - Opposition                   & 0.234  &  0.0533     & 0.249 &   0.7213    & 0.272   & 0.0852  
\\
\hline
\end{tabular}
}
\caption{\textbf{Correlation coefficients and p-values for each pair of labels in the dataset.}}
\label{tab:corr_coeff}
\end{table}

\begin{figure*}
\subfloat[Allegation]{%
  \includegraphics[width=0.25\textwidth]{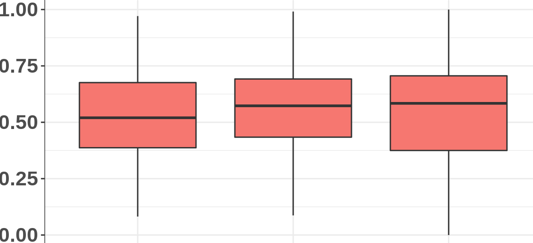}
}
\subfloat[Directed Hate]{%
  \includegraphics[width=0.25\textwidth]{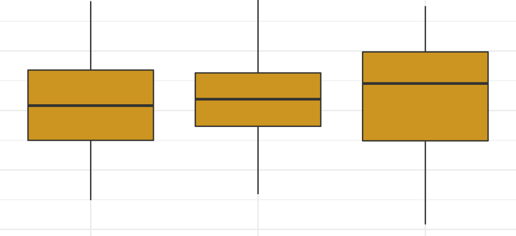}
}
\subfloat[Generalized Hate]{%
  \includegraphics[width=0.25\textwidth]{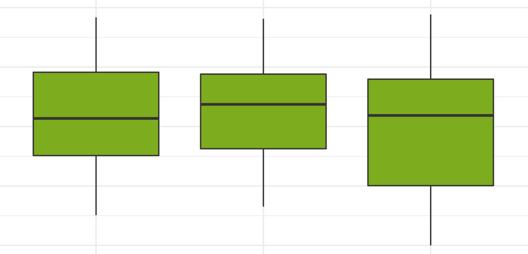}
}
\subfloat[Opposition]{%
  \includegraphics[width=0.25\textwidth]{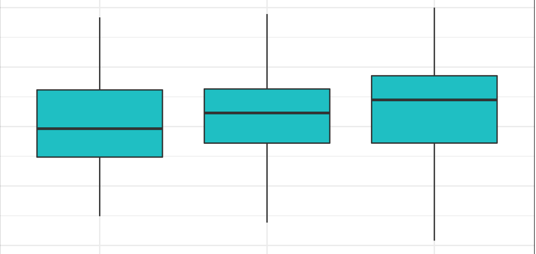}
}
\newline
\subfloat[Justification]{%
  \includegraphics[width=0.25\textwidth]{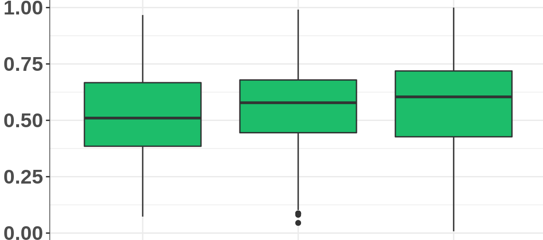}
}
\subfloat[Sarcasm]{%
  \includegraphics[width=0.25\textwidth]{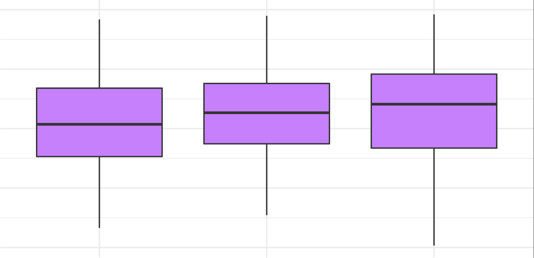}
}
\subfloat[Refutation]{%
  \includegraphics[width=0.25\textwidth]{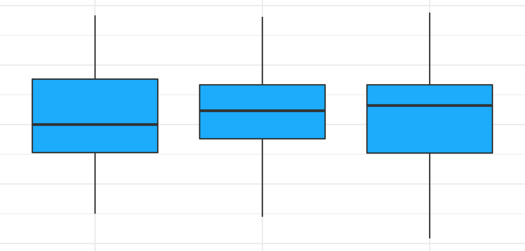}
}
\subfloat[Support]{%
  \includegraphics[width=0.25\textwidth]{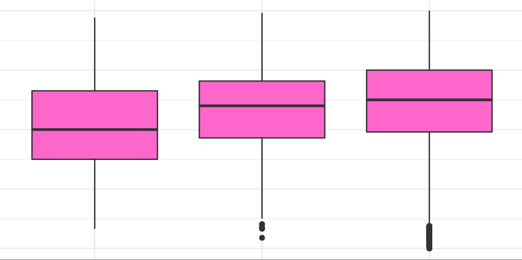}
}
\caption{\textbf{Arousal, Dominance, and Valence scores for all class labels based on NRC VAD lexicon for each of the labels. The first box presents arousal score, the second one dominance score, and the third one valence dimension.}}
\label{fig:nrc_vad}
\end{figure*}

\subsection{Keywords}
We used SAGE \cite{eisenstein2011sparse}, a topic modelling method, to identify keywords associated with the various class labels in our dataset. SAGE is an unsupervised generative model that can identify words that distinguish one part of the corpus from rest. For our keyword analysis, we removed all the hashtags and only considered tokens that appeared at least five times in the corpus, thus ensuring they were representative of the topic. Table \ref{tab:sage_res} presents the top five keywords associated with each class and also their salience scores. Though \textit{Directed} and \textit{Generalized} hate are closely related topics, there is not much overlap between the top 5 salient keywords suggesting that there are linguistic cues to distinguish between them. The word {\fontfamily{qcr}\selectfont predators} is strongly indicative of \textit{Generalized Hate}, which is intuitive because it is a term often used to describe people who were accused of sexual misconduct. The word {\fontfamily{qcr}\selectfont lol} being associated with \textit{Sarcasm} is also reasonably intuitive because of sarcasm's close relation with humour.

\subsection{Sentiment Analysis}
Figure \ref{fig:nrc_sentiment} presents a word cloud representation of the data where the colours are assigned based on NRC emotion lexicon \cite{Mohammad13}: green for positive and red for negative. We also analyzed all the classes in terms of Valence, Arousal, and Dominance using the NRC VAD lexicon \cite{vad-acl2018}. The results are summarized in Figure \ref{fig:nrc_vad}. Of all the classes, \textit{Directed-Hate} has the largest valence spread, which is likely because of the extreme nature of the opinions expressed in such tweets. The spread for the dominance is fairly narrow for all class labels with the median score slightly above 0.5, suggesting a slightly dominant nature exhibited by the authors of the tweets.

\section*{Discussion}
\label{sec:discuss}
This paper introduces a new dataset containing tweets related to the \#MeToo movement. It may involve opinions over socially stigmatized issues or self-reports of distressing incidents. Therefore, it is necessary to examine the social impact of this exercise, the ethics of the individuals concerned with the dataset, and it's limitations. 

\noindent\textbf{Mental health implications: }
This dataset open sources posts curated by individuals who may have undergone instances of sexual exploitation in the past. While we respect and applaud their decision to raise their voices against their exploitation, we also understand that their revelations may have been met with public backlash and apathy in both the virtual as well as the real world. In such situations, where the social reputation of both accuser and accused may be under threat, mental health concerns become very important\footnote{https://www.thelancet.com/journals/lancet/article/PIIS0140-6736(18)30991-7/fulltext}. As survivors recount their horrific episodes of sexual harassment, it becomes imperative to provide them with therapeutic care \cite{fredriksen2014creating} as a safeguard against mental health hazards. Such measures, if combined with the integration of mental health assessment tools in social media platforms, can make victims of sexual abuse feel more empowered and self-contemplative towards their revelations.\\

\noindent\textbf{Use of MeTooMA dataset for population studies: } We would like to mention that there have been no attempts to conduct population-centric analysis on the proposed dataset. The analysis presented in this dataset should be seen as a proof of concept to examine the instances of \#MeToo movement on Twitter. The authors acknowledge that learning from this dataset cannot be used as-is for any direct social interventions. Network sampling of real-world users for any experimental work beyond this dataset would require careful evaluation beyond the observational analysis presented herein. Moreover, the findings could be used to assist already existing human knowledge. Experiences of the affected communities should be recorded and analyzed carefully, which could otherwise lead to social stigmatization, discrimination and societal bias. Enough care has been ensured so that this work does not come across as trying to target any specific individual for their personal stance on the issues pertaining to the social theme at hand. The authors do not aim to vilify individuals accused in the \#MeToo cases in any manner. Our work tries to bring out general trends that may help researchers develop better techniques to understand mass unorganized virtual movements.\\

\noindent\textbf{Effect on marginalized communities: } The authors recognize the impact of the \#MeToo movement on socially stigmatized populations like LGBTQIA+. The \#MeToo movement provided such individuals with the liberty to express their notions about instances of sexual violence and harassment\footnote{https://rewire.news/article/2018/10/09/for-lgbtq-youth-metoo-is-not-a-heteronormative-issue/}. The movement acted as a catalyst towards implementing social policy changes to benefit the members of these communities\footnote{https://www.reuters.com/article/us-lgbt-rights-twitter/mequeer-takes-twitter-by-storm-as-lgbt-community-cries-metoo-idUSKCN1L71WW}. Hence, it is essential to keep in mind that any experimental work undertaken on this dataset should try to minimize the biases against the minority groups which might get amplified in cases of sudden outburst of public reactions over sensitive media discussions. \\ 

\noindent\textbf{Limitations of individual consent: }Considering the mental health aspects of the individuals concerned, social media practitioners should vary of making automated interventions to aid the victims of sexual abuse as some individuals might not prefer to disclose their sexual identities or notions. Concerned social media users might also repeal their social media information if found out that their personal information may be potentially utilised for computational analysis. Hence, it is imperative to seek subtle individual consent before trying to profile authors involved in online discussions to uphold personal privacy. \\
\section*{Use Cases}
The authors would like to formally propose some ideas on possible extensions of the proposed dataset:

\begin{itemize}
    \item The rise of online \textbf{hate speech} and its related behaviours like cyber-bullying has been a hot topic of research in gender studies \cite{djuric2015hate}. Our dataset could be utilized for extracting actionable insights and virtual dynamics to identify gender roles for analyzing sexual abuse revelations similar to \cite{yuce2014bridging}.
    
    \item The dataset could be utilized by psycholinguistics for extracting contextualized lexicons to examine how influential people are portrayed on public platforms in events of mass social media movements \cite{field2019contextual}. Interestingly, such analysis may help linguists determine the \textbf{power dynamics of authoritative people} in terms of perspective and sentiment through campaign modelling.
    
    \item Marginalized voices affected by mass social movements can be studied through \textbf{polarization analysis} on graph-based simulations of the social media networks. Based on the data gathered from these nodes, community interactions could be leveraged to identify indigenous issues pertaining to societal unrest across various sections of the society\cite{rho2018fostering}.
    
    \item \textbf{Challenge Proposal}: The authors of the paper would like to extend the present work as a challenge proposal for building computational semantic analysis systems aimed at online social movements. In contrast to already available datasets and existing challenges, we propose tasks on detecting hate speech, sarcasm, stance and relevancy that will be more focused on social media activities surrounding revelations of sexual abuse and harassment. The tasks may utilize the message-level text, linked images, tweet-level metadata and user-level interactions to model systems that are \textbf{F}air, \textbf{A}ccountable, \textbf{I}nterpretable and \textbf{R}esponsible (FAIR).
    
\end{itemize}
Research ideas emerging from this work should not be limited to the above discussion. If needed, supplementary data required to enrich this dataset can be collected utilizing Twitter API and \textit{JSON} records for exploratory tasks beyond the scope of the paper.

\section*{Conclusion}
In this paper, we presented a new dataset annotated for five different linguistic aspects: relevance, stance, hate speech, sarcasm, and dialogue acts. To our knowledge, there are no datasets out there that provide annotations across so many different dimensions. This allows researchers to perform various multi-label and multi-aspect classification experiments. Additionally, researchers could also address some interesting questions on how different linguistic components influence each other: e.g. does understanding one's stance help in better prediction of hate speech? 

In addition to these exciting computational challenges, we expect this data could be useful for socio and psycholinguists in understanding the language used by victims when disclosing their experiences of abuse. Likewise, they could analyze the language used by alleged individuals in justifying their actions. It also provides a chance to examine the language used to express hate in the context of sexual abuse.

In the future, we would like to propose challenge tasks around this data where the participants will have to build computational models to capture all the different linguistic aspects that were annotated. We expect such a task would drive researchers to ask more interesting questions, find limitations of the dataset, propose improvements, and provide interesting insights.

\bibliographystyle{aaai}
\bibliography{ref.bib}

\begin{thebibliography}{}

\bibitem[\protect\citeauthoryear{Ang, Liu, and
  Shriberg}{2005}]{ang2005automatic}
Ang, J.; Liu, Y.; and Shriberg, E.
\newblock 2005.
\newblock Automatic dialog act segmentation and classification in multiparty
  meetings.
\newblock In {\em Proceedings.(ICASSP'05). IEEE International Conference on
  Acoustics, Speech, and Signal Processing, 2005.}, volume~1,  I--1061.
\newblock IEEE.

\bibitem[\protect\citeauthoryear{Artstein and Poesio}{2008}]{artstein2008inter}
Artstein, R., and Poesio, M.
\newblock 2008.
\newblock Inter-coder agreement for computational linguistics.
\newblock {\em Computational Linguistics} 34(4):555--596.

\bibitem[\protect\citeauthoryear{Augenstein \bgroup et al\mbox.\egroup
  }{2016}]{augenstein2016stance}
Augenstein, I.; Rockt{\"a}schel, T.; Vlachos, A.; and Bontcheva, K.
\newblock 2016.
\newblock Stance detection with bidirectional conditional encoding.
\newblock {\em arXiv preprint arXiv:1606.05464}.

\bibitem[\protect\citeauthoryear{Badjatiya \bgroup et al\mbox.\egroup
  }{2017}]{badjatiya2017deep}
Badjatiya, P.; Gupta, S.; Gupta, M.; and Varma, V.
\newblock 2017.
\newblock Deep learning for hate speech detection in tweets.
\newblock In {\em Proceedings of the 26th International Conference on World
  Wide Web Companion},  759--760.
\newblock International World Wide Web Conferences Steering Committee.

\bibitem[\protect\citeauthoryear{Bamman and
  Smith}{2015}]{bamman2015contextualized}
Bamman, D., and Smith, N.~A.
\newblock 2015.
\newblock Contextualized sarcasm detection on twitter.
\newblock In {\em Ninth International AAAI Conference on Web and Social Media}.

\bibitem[\protect\citeauthoryear{Basile \bgroup et al\mbox.\egroup
  }{2019}]{basile-etal-2019-semeval}
Basile, V.; Bosco, C.; Fersini, E.; Nozza, D.; Patti, V.; Rangel~Pardo, F.~M.;
  Rosso, P.; and Sanguinetti, M.
\newblock 2019.
\newblock {S}em{E}val-2019 task 5: Multilingual detection of hate speech
  against immigrants and women in twitter.
\newblock In {\em Proceedings of the 13th International Workshop on Semantic
  Evaluation},  54--63.
\newblock Minneapolis, Minnesota, USA: Association for Computational
  Linguistics.

\bibitem[\protect\citeauthoryear{Chowdhury \bgroup et al\mbox.\egroup
  }{2019}]{chowdhury2019speak}
Chowdhury, A.~G.; Sawhney, R.; Mathur, P.; Mahata, D.; and Shah, R.~R.
\newblock 2019.
\newblock Speak up, fight back! detection of social media disclosures of sexual
  harassment.
\newblock In {\em Proceedings of the 2019 Conference of the North American
  Chapter of the Association for Computational Linguistics: Student Research
  Workshop},  136--146.

\bibitem[\protect\citeauthoryear{Djuric \bgroup et al\mbox.\egroup
  }{2015}]{djuric2015hate}
Djuric, N.; Zhou, J.; Morris, R.; Grbovic, M.; Radosavljevic, V.; and
  Bhamidipati, N.
\newblock 2015.
\newblock Hate speech detection with comment embeddings.
\newblock In {\em Proceedings of the 24th international conference on world
  wide web},  29--30.
\newblock ACM.

\bibitem[\protect\citeauthoryear{Eisenstein, Ahmed, and
  P~Xing}{2011}]{eisenstein2011sparse}
Eisenstein, J.; Ahmed, A.; and P~Xing, E.
\newblock 2011.
\newblock Sparse additive generative models of text.

\bibitem[\protect\citeauthoryear{Field, Bhat, and
  Tsvetkov}{2019}]{field2019contextual}
Field, A.; Bhat, G.; and Tsvetkov, Y.
\newblock 2019.
\newblock Contextual affective analysis: A case study of people portrayals in
  online\# metoo stories.
\newblock In {\em Proceedings of the International AAAI Conference on Web and
  Social Media}, volume~13,  158--169.

\bibitem[\protect\citeauthoryear{Fredriksen-Goldsen \bgroup et al\mbox.\egroup
  }{2014}]{fredriksen2014creating}
Fredriksen-Goldsen, K.~I.; Hoy-Ellis, C.~P.; Goldsen, J.; Emlet, C.~A.; and
  Hooyman, N.~R.
\newblock 2014.
\newblock Creating a vision for the future: Key competencies and strategies for
  culturally competent practice with lesbian, gay, bisexual, and transgender
  (lgbt) older adults in the health and human services.
\newblock {\em Journal of gerontological social work} 57(2-4):80--107.

\bibitem[\protect\citeauthoryear{Hosterman \bgroup et al\mbox.\egroup
  }{2018}]{hosterman2018twitter}
Hosterman, A.~R.; Johnson, N.~R.; Stouffer, R.; and Herring, S.
\newblock 2018.
\newblock Twitter, social support messages, and the\# metoo movement.
\newblock {\em The Journal of Social Media in Society} 7(2):69--91.

\bibitem[\protect\citeauthoryear{Hutchings}{2012}]{hutchings2012commercial}
Hutchings, C.
\newblock 2012.
\newblock Commercial use of facebook and twitter--risks and rewards.
\newblock {\em Computer Fraud \& Security} 2012(6):19--20.

\bibitem[\protect\citeauthoryear{Khatua, Cambria, and
  Khatua}{2018}]{khatua2018sounds}
Khatua, A.; Cambria, E.; and Khatua, A.
\newblock 2018.
\newblock Sounds of silence breakers: Exploring sexual violence on twitter.
\newblock In {\em 2018 IEEE/ACM International Conference on Advances in Social
  Networks Analysis and Mining (ASONAM)},  397--400.
\newblock IEEE.

\bibitem[\protect\citeauthoryear{Kim, Cavedon, and
  Baldwin}{2010}]{kim2010classifying}
Kim, S.~N.; Cavedon, L.; and Baldwin, T.
\newblock 2010.
\newblock Classifying dialogue acts in one-on-one live chats.
\newblock In {\em Proceedings of the 2010 Conference on Empirical Methods in
  Natural Language Processing},  862--871.
\newblock Association for Computational Linguistics.

\bibitem[\protect\citeauthoryear{Krippendorff}{2011}]{krippendorff2011computing}
Krippendorff, K.
\newblock 2011.
\newblock Computing krippendorff's alpha-reliability.

\bibitem[\protect\citeauthoryear{Lopez, Muldoon, and
  McKeown}{2019}]{lopez2019one}
Lopez, K.~J.; Muldoon, M.~L.; and McKeown, J.~K.
\newblock 2019.
\newblock One day of\# feminism: Twitter as a complex digital arena for
  wielding, shielding, and trolling talk on feminism.
\newblock {\em Leisure Sciences} 41(3):203--220.

\bibitem[\protect\citeauthoryear{McTear, Callejas, and
  Griol}{2016}]{mctear2016conversational}
McTear, M.~F.; Callejas, Z.; and Griol, D.
\newblock 2016.
\newblock {\em The conversational interface}, volume~6.
\newblock Springer.

\bibitem[\protect\citeauthoryear{Mohammad and Turney}{2013}]{Mohammad13}
Mohammad, S.~M., and Turney, P.~D.
\newblock 2013.
\newblock Crowdsourcing a word-emotion association lexicon.
\newblock 29(3):436--465.

\bibitem[\protect\citeauthoryear{Mohammad \bgroup et al\mbox.\egroup
  }{2016}]{mohammad2016semeval}
Mohammad, S.; Kiritchenko, S.; Sobhani, P.; Zhu, X.; and Cherry, C.
\newblock 2016.
\newblock Semeval-2016 task 6: Detecting stance in tweets.
\newblock In {\em Proceedings of the 10th International Workshop on Semantic
  Evaluation (SemEval-2016)},  31--41.

\bibitem[\protect\citeauthoryear{Mohammad}{2018}]{vad-acl2018}
Mohammad, S.~M.
\newblock 2018.
\newblock Obtaining reliable human ratings of valence, arousal, and dominance
  for 20,000 english words.
\newblock In {\em Proceedings of The Annual Conference of the Association for
  Computational Linguistics (ACL)}.

\bibitem[\protect\citeauthoryear{Pandey \bgroup et al\mbox.\egroup
  }{2018}]{pandey2018distributional}
Pandey, R.; Purohit, H.; Stabile, B.; and Grant, A.
\newblock 2018.
\newblock Distributional semantics approach to detect intent in twitter
  conversations on sexual assaults.
\newblock In {\em 2018 IEEE/WIC/ACM International Conference on Web
  Intelligence (WI)},  270--277.
\newblock IEEE.

\bibitem[\protect\citeauthoryear{Rajadesingan, Zafarani, and
  Liu}{2015}]{rajadesingan2015sarcasm}
Rajadesingan, A.; Zafarani, R.; and Liu, H.
\newblock 2015.
\newblock Sarcasm detection on twitter: A behavioral modeling approach.
\newblock In {\em Proceedings of the Eighth ACM International Conference on Web
  Search and Data Mining},  97--106.
\newblock ACM.

\bibitem[\protect\citeauthoryear{Rho, Mark, and
  Mazmanian}{2018}]{rho2018fostering}
Rho, E. H.~R.; Mark, G.; and Mazmanian, M.
\newblock 2018.
\newblock Fostering civil discourse online: Linguistic behavior in comments
  of\# metoo articles across political perspectives.
\newblock {\em Proceedings of the ACM on Human-Computer Interaction}
  2(CSCW):147.

\bibitem[\protect\citeauthoryear{Schrading \bgroup et al\mbox.\egroup
  }{2015}]{schrading2015analysis}
Schrading, N.; Alm, C.~O.; Ptucha, R.; and Homan, C.
\newblock 2015.
\newblock An analysis of domestic abuse discourse on reddit.
\newblock In {\em Proceedings of the 2015 Conference on Empirical Methods in
  Natural Language Processing},  2577--2583.

\bibitem[\protect\citeauthoryear{Sharifirad and
  Jacovi}{2019}]{sharifirad2019learning}
Sharifirad, S., and Jacovi, A.
\newblock 2019.
\newblock Learning and understanding different categories of sexism using
  convolutional neural network’s filters.
\newblock In {\em Proceedings of the 2019 Workshop on Widening NLP},  21--23.

\bibitem[\protect\citeauthoryear{Sobhani}{2017}]{sobhani2017stance}
Sobhani, P.
\newblock 2017.
\newblock {\em Stance detection and analysis in social media}.
\newblock Ph.D. Dissertation, Universit{\'e} d'Ottawa/University of Ottawa.

\bibitem[\protect\citeauthoryear{Waseem and Hovy}{2016}]{waseem2016hateful}
Waseem, Z., and Hovy, D.
\newblock 2016.
\newblock Hateful symbols or hateful people? predictive features for hate
  speech detection on twitter.
\newblock In {\em Proceedings of the NAACL student research workshop},  88--93.

\bibitem[\protect\citeauthoryear{Yuce \bgroup et al\mbox.\egroup
  }{2014}]{yuce2014bridging}
Yuce, S.~T.; Agarwal, N.; Wigand, R.~T.; Lim, M.; and Robinson, R.~S.
\newblock 2014.
\newblock Bridging women rights networks: Analyzing interconnected online
  collective actions.
\newblock {\em Journal of Global Information Management (JGIM)} 22(4):1--20.

\bibitem[\protect\citeauthoryear{Zarisheva and
  Scheffler}{2015}]{zarisheva2015dialog}
Zarisheva, E., and Scheffler, T.
\newblock 2015.
\newblock Dialog act annotation for twitter conversations.
\newblock In {\em Proceedings of the 16th Annual Meeting of the Special
  Interest Group on Discourse and Dialogue},  114--123.

\end{thebibliography}

\end{document}